\documentclass[10pt,twocolumn,letterpaper]{article}

\usepackage{iccv}
\usepackage{times}
\usepackage{epsfig}
\usepackage{graphicx}
\usepackage{amsmath}
\usepackage{amssymb}
\usepackage{booktabs}
\usepackage{multicol}
\usepackage{multirow}
\newcommand{\vPhi}{{\mathbf{\Phi}}}
\newcommand{\vh}{{\mathbf{h}}}
\newcommand{\vu}{{\mathbf{u}}}
\newcommand{\vv}{{\mathbf{v}}}
\newcommand{\vx}{{\mathbf{x}}}
\newcommand{\vzero}{{\mathbf{0}}}
\newcommand{\CA}{{\mathcal{A}}}
\newcommand{\CH}{{\mathcal{H}}}
\newcommand{\CL}{{\mathcal{L}}}
\newcommand{\CP}{{\mathcal{P}}}
\newcommand{\CN}{{\mathcal{N}}}
\newcommand{\DD}{{\mathbb{D}}}
\newcommand{\myparagraph}[1]{\vspace{3pt}\noindent{\bf #1}}
\usepackage[pagebackref=true,breaklinks=true,letterpaper=true,colorlinks,bookmarks=false]{hyperref}

\iccvfinalcopy % *** Uncomment this line for the final submission

\def\iccvPaperID{6834} % *** Enter the ICCV Paper ID here

\ificcvfinal\fi

\begin{document}

%%%%%%%%% TITLE
\title{Hierarchical Explanations for Video Action Recognition}

\author{Sadaf Gulshad,
\quad Teng Long, 
\quad Nanne van Noord \\
University of Amsterdam\\
{\tt\small \{s.gulshad, t.long, n.j.e.vannoord\}@uva.nl}
}

\maketitle
% Remove page # from the first page of camera-ready.
\ificcvfinal\thispagestyle{empty}\fi

%%%%%%%%% ABSTRACT
\begin{abstract}
To interpret deep neural networks, one main approach is to dissect the visual input and find the prototypical parts responsible for the classification. However, existing methods often ignore the hierarchical relationship between these prototypes, and thus can not explain semantic concepts at both higher level (e.g., water sports) and lower level (e.g., swimming). In this paper inspired by human cognition system, we leverage hierarchal information to deal with uncertainty: When we observe water and human activity, but no definitive action it can be recognized as the water sports parent class. Only after observing a person swimming can we definitively refine it to the swimming action. To this end, we propose HIerarchical Prototype Explainer (HIPE) to build hierarchical relations between prototypes and classes. HIPE enables a reasoning process for video action classification by dissecting the input video frames on multiple levels of the class hierarchy, our method is also applicable to other video tasks.  The faithfulness of our method is verified by reducing accuracy-explainability trade off on ActivityNet and UCF-101 while providing multi-level explanations.
\end{abstract}

%%%%%%%%% BODY TEXT

\section{Introduction}
\label{sec:intro}
\begin{figure}
    \centering
    \includegraphics[width=\linewidth]{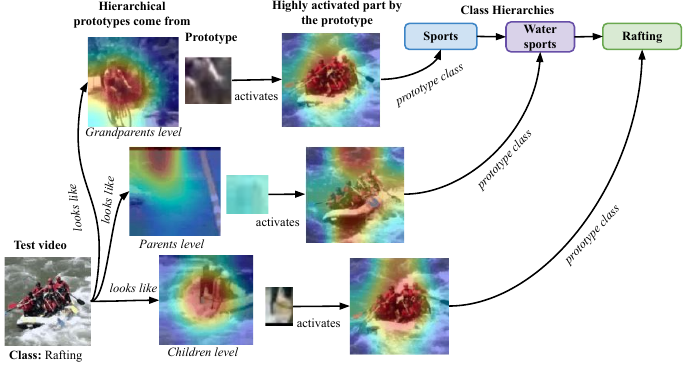}
    \caption{Our approach generates explanations by learning hierarchical prototypes at different levels: at grandparent level prototypes belonging to sports class, parent level belonging to water sports class, and class level belonging to rafting class. }
    \label{fig:Hyperbole}
    % \vspace{-2mm}
\end{figure}

When describing the world around us we may do so at different levels of granularity, depending on the information available or the level of detail we intend to convey.
For instance, a video might open with a shot of a cheering crowd, allowing us to recognize it as a \textit{a sports event}, as the camera then pans to the river we can deduce that it is \textit{a water sports event}. However, only when the raft comes into the frame can we determine that it concerns \textit{rafting}. Nonetheless, in our description of this video, we may still only refer to it as a sports or water sports event. Our reasoning and description processes build on the hierarchical relation between classes, allowing for navigation between generic and specific. In this work, we implement this process for video action recognition by learning hierarchical concepts that we leverage for improved classification performance and explanations at multiple levels of granularity, Figure \ref{fig:Hyperbole}.
% explainable AI-> for videos
\par
Despite the remarkable performance of neural networks for video understanding tasks \cite{hara3dcnns,KarpathyCVPR14,peng2016multi,singh2017online,simonyan2014two,wang2016temporal,carreira2017quo,bertinetto2016fully} it is still hard to explain the decisions of these networks, which is of utmost importance for practical application. This necessity has led to a growing stream of research that focuses on making models interpretable besides performing accurately \cite{Hendricks_2018_ECCV, Kim_2018_ECCV,10.1145/3372278.3390672,trinh2021interpretable, li2021towards}. A promising line of posthoc explainability methods are the concept bottleneck models \cite{koh2020concept, losch2019interpretability, kim2018interpretability}, which focus on explaining the decisions of neural networks by predicting human understandable concepts before performing final prediction. However, these methods require dense concept annotations or access from other resources like pre-trained language models \cite{yuksekgonul2022post, oikarinenlabel}. Our method for providing builtin explanations does not require dense annotations.

In the similar spirit prototype-based models \cite{li2018deep, chen2019looks,donnelly2022deformable} focus on learning prototypes during training and make predictions based on the learned prototypes during inference. This enables \textit{this look like that} explanations. However, previous case-based reasoning works are limited to 2D images and models. Moreover, they provide a single level of explanations and in case of uncertainty, the explanations can be as bad as arbitrary, as each explanation is considered equally apart. In this work, we focus on capturing the hierarchical relations between actions to provide multi-level explanations for videos.     

A challenge for explainable models, such as concept bottleneck models and prototype-based models, is that it introduces an accuracy-explainability trade off, where explainability comes at the cost of accuracy. With this paper, we aim to introduce a model with built-in explainability which is less affected by this trade off. To achieve this goal we are inspired by recent works on learning hyperbolic embedding spaces, as opposed to euclidean, in natural language processing \cite{tifrea2018poincar,dhingra2018embedding} and computer vision tasks \cite{atigh2022hyperbolic,ghadimi2021hyperbolic,long2020searching}. These works have demonstrated that it is beneficial for performance to have the embedding space be guided by hierarchical knowledge, which we believe will also benefit explainability.
This belief is guided by the hierarchical cognition
process of humans, that is we are likely to organize concepts from specific to general \cite{warrington1975selective, minsky1982semantic, mcclelland2003parallel}, and the representation of categories in the hyperbolic space. 

The main contributions of our paper are: 1) We propose Hierarchical Prototype Explainer (HIPE), a reasoning model for interpreting video action recognition. 2) We demonstrate that HIPE can provide meaningful explanations even in the case of uncertainty or lack of information by providing multi-level explanations i.e., at class, parent, or grandparent level. 3) We perform a benchmark and show that HIPE counters accuracy-explainability trade off.     

\section{Related Work}
\subsection{Interpretations for Videos}
Interpretations for neural networks can be broadly classified into two categories: 1) fitting explanations to the decisions of the network after it has been trained i.e. \textit{posthoc} \cite{Hendricks_2018_ECCV,Kim_2018_ECCV,10.1145/3372278.3390672,ribeiro2016should,lundberg2017unified}, 2) building explanation mechanism inherent in the network i.e. \textit{built-in} explanations.  
\cite{trinh2021interpretable, li2021towards, lin2021gradient, uchiyama2022visually}. In this work, we focus on learning semantic representations which are used for classification during training rather than explaining a black box network posthoc. 

A great deal of previous work has focused on video action recognition, detection, segmentation and more \cite{hara3dcnns,KarpathyCVPR14,peng2016multi,singh2017online,simonyan2014two,wang2016temporal,carreira2017quo,bertinetto2016fully}, however, most of these works focus on designing black box models for specific tasks. They do not explain why a certain decision is made by the model. Moreover, most of the research in the domain of visual explanations focuses on images. Only a few works focus on the interpretation of these networks for videos \cite{karpathy2015visualizing, bargal2018excitation, stergiou2019saliency, li2021towards, stergiou2019class}, and it is not possible to directly apply image-based explanation methods to videos due to an extra time dimension in videos.   

\cite{karpathy2015visualizing} and \cite{bargal2018excitation} focus on visualizing spatio-temporal attention in RNNs, CNNs are used only to extract features. Inspired by class activation maps (CAM) \cite{zhou2016learning} for images \cite{stergiou2019saliency} extended it for videos by finding both regions and frames responsible for classification. \cite{li2021towards} utilized perturbations to extract the most informative parts of the inputs responsible for the outputs. Both \cite{stergiou2019saliency, li2021towards} are posthoc methods, which means they do not use explanations during prediction therefore they might not be faithful to what the network computes \cite{donnelly2022deformable}.   \cite{stergiou2019class} introduced class feature pyramids, a method that traverses through the whole network and searches for the kernels at different depths of the network responsible for classification, therefore this method is computationally expensive. In contrast, we enable built-in multi-level explanations that do not add any computational complexity.  
In this paper, we enable multi-level explanations for videos by learning hierarchical prototypes for each class and tracing them back to input videos. 

\subsection{Case-based Reasoning Models}
There are two main categories of case-based reasoning models: \textit{concept bottleneck models} which introduce a bottleneck layer that learns human understandable concepts, and \textit{prototype-based models} that learn prototypes that are closer to the samples in the training set. Concept bottleneck models provide posthoc explanations by replacing the final layer of the neural network with a layer that predicts human understandable concepts \eg for a cardinal bird class the concepts will be red wings, red beak, black eye \cite{koh2020concept, losch2019interpretability, kim2018interpretability, wang2022hint}. These predicted concepts are then used to perform classification. However, concept bottleneck models require dense concept annotations for the model to learn them. Moreover, as they use predicted concepts to perform classification therefore, they suffer from explainability-accuracy trade off. \cite{yuksekgonul2022post, oikarinenlabel} focus on addressing these limitations by either incorporating concepts by  transferring them from natural language descriptions or generating them from a GPT model. In contrast, our work focuses on providing built-in explanations by learning representative samples for each class and its (grand)parent class, while countering explainability-accuracy trade off. Our method do not require heavy annotations but utilize either the hierarchy available with the datasets or it can be easily defined based on the relations between classes. 

More relevant to our method, \cite{wang2022hint} build relations between neurons and hierarchical concepts. However, our method differs from \cite{wang2022hint} in three significant ways 1) their method is posthoc (a separate concept classifier is trained) while ours is builtin, 2) to learn hierarchical concepts they utilize wordnet hierarchy of Imagenet while we do not, 3) considers the contribution of neurons (non comprehensible for humans), whereas we visualize learned features by the prototype layer.

% In machine learning the term ``prototype'' is used in various contexts, in zero-shot learning \cite{xu2022attribute, xu2020zeroshot, li2017zero, yu2020episode} and few-shot learning \cite{snell2017prototypical, pahde2021multimodal} prototypes are points in the embedding space representing a single class and the distance from these prototypes is used during classification. However, in our work we learn protoypes that are closer to the samples in the training set, and multiple prototypes are used to represent each class. They are optimized to resemble the training set in order to provide visual explanations.  

The idea behind prototype-based models, to provide built-in explanations with prototypes was first explored in \cite{li2018deep}, where the authors introduced a prototype layer in the network with an encoder-decoder architecture. The prototype layer stores weights which are close to encoded training samples, and a decoder is used to visualize them. However, their model fails to generate realistic visualizations for natural images. Thus \cite{chen2019looks} proposed to learn prototypes for each class and visualized them by tracing them back to the input images without a decoder. We get inspiration from \cite{chen2019looks} to provide built-in explanations, but where their work is limited to 2D images and provides only one-level explanations we extend it to multi-level explanations for videos.

% \cite{nauta2021looks} enhances the prototypical explanations by adding textual explanations to explain prototypes. 

\cite{rymarczyk2020protopshare} focuses on reducing the number of prototypes for each class by finding shared prototypes among classes.
\cite{wang2021interpretable} introduced a different similarity metric for computing similarities between prototypes and image patches.  Deformable ProtoPNet \cite{donnelly2022deformable} learns spatially flexible prototypes to capture pose and context variations in the input. All previous prototype-based explanation methods provide explanations without considering the hierarchical relations between classes on well-defined image CUB birds \cite{wah2011caltech} and Stanford cars \cite{KrauseStarkDengFei-Fei_3DRR2013} datasets.  In  
contrast inspired by the human way of explanations we consider hierarchical relations between classes while learning prototypes for each class for video datasets. 

% \cite{nauta2021neural} introduced neural PrototypeTrees by combining CNN architecture with the soft binary trees for providing local and global explanations.

Most closely related to our work are \cite{trinh2021interpretable,nauta2021neural}. \cite{trinh2021interpretable} introduced a dynamic prototype network (DPNet) for finding temporal artifacts and unnatural movements in deep fake videos. However, the goal of DPNet is different from ours with only two target classes fake/real making the task easier. PrototypeTrees \cite{nauta2021neural} use a decision-tree with a pre-defined structure, where individual prototypes are learned at each decision. The prototypes are optimised to increase purity along the path through the tree. However, for PrototypeTrees the position in, and order of, the tree does not describe a hierarchy, that is closer to the root does not imply a more general semantic level. Moreover, as the number of prototypes depends upon the size of the tree, learning a ProtoTree becomes computationally expensive. Our proposed multi-level explanations follow a very clear semantic distribution , where (grand)parent prototypes are more generic and do not add any extra computational complexity.
\begin{figure}
    \centering
    \includegraphics[width=\linewidth]{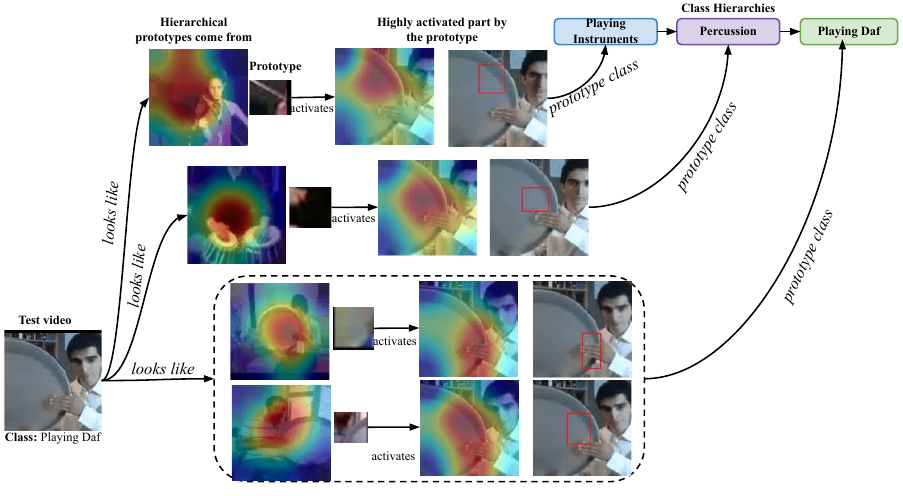}
    \caption{\textbf{Significance of Hierarchical Explanations.} Hierarchical explanations generated for an input video of playing daf. Even if someone does not know what is \textit{``playing daf''}, its parent explanation shows that it involves \textit{``percussion''} and grandparent shows that it is a musical \textit{``instrument''}. }
    \label{fig:results-1}
    \vspace{-3mm}
\end{figure}

\begin{figure*}
    \centering
    \includegraphics[width=\linewidth]{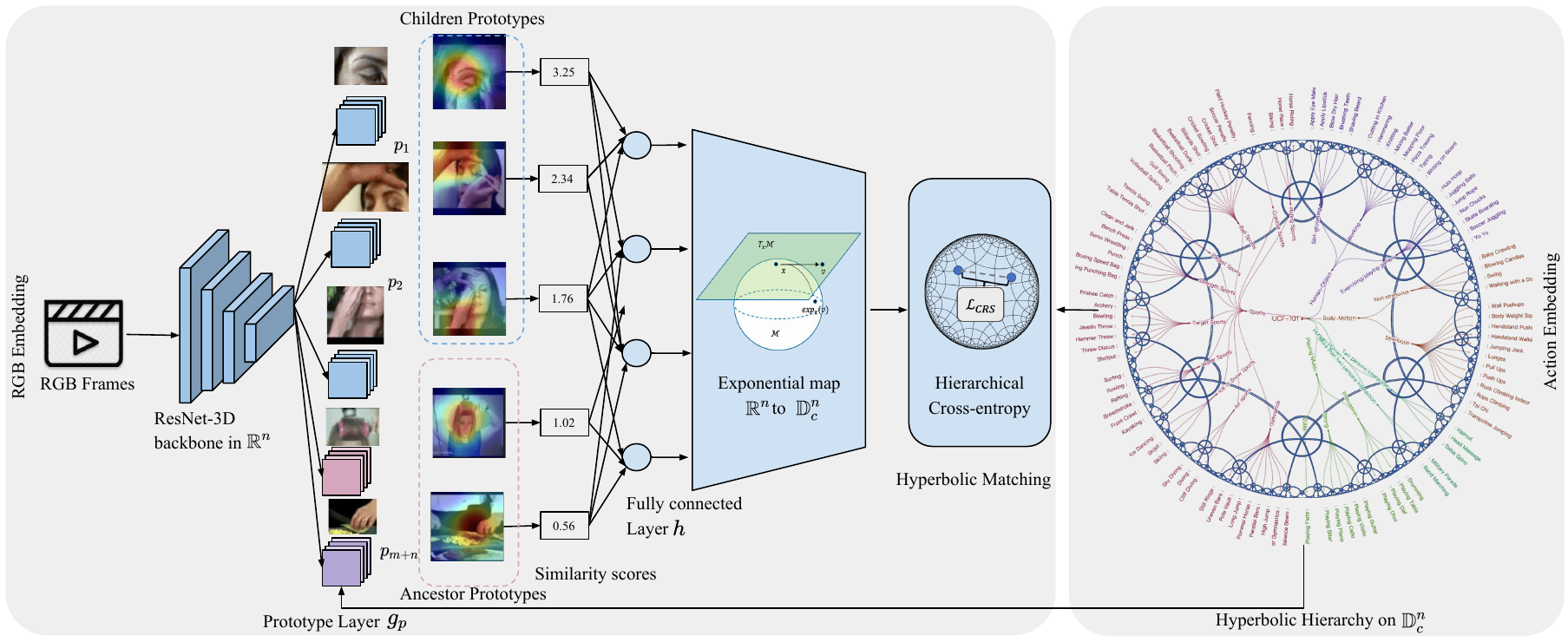}
    \caption{\textbf{Overview of the Hierarchical Prototype Explainer.} The Resnet-3D backbone extracts video features and the prototype layer learns prototypes for children and parents, these prototypes are then converted to a single similarity score through max pooling. Finally, scores are converted from $\mathbb{R}^n$ to $\DD^n$ through a fully connected layer followed by an exponential map, to the shared hyperbolic space for hierarchical learning. Actions are mapped onto the shared hyperbolic space by learning a discriminative embedding on $\DD^n$.}
    \label{fig:Architecture}
    
\end{figure*}

\subsection{Hyperbolic Embeddings}
Hyperbolic embeddings have recently received increased attention as they enable {continuous} representations of hierarchical knowledge \cite{nickel2017poincare,chami2019hyperbolic}. This continuous nature makes it such that information of (grand)parent classes is implicitly included, allowing hyperbolic training to remain single-label. Their effectiveness has also been shown for textual \cite{tifrea2018poincar,ganea2018hyperbolic, zhu2020hypertext} and visual data \cite{khrulkov2020hyperbolic,atigh2022hyperbolic,ghadimi2021hyperbolic,long2020searching}. Hyperbolic embeddings have also been used for zero-shot learning \cite{liu2020hyperbolic,fang2021kernel} and for video action recognition \cite{long2020searching,suris2021learning}. The hierarchical relationship between videos and the hierarchical way of explaining decisions for humans calls for the need of using hyperbolic spaces. Here, we utilize hyperbolic embeddings for learning hierarchical prototypes to provide human-like explanations for video action recognition.

\section{Hierarchical Explanations}
Single level explanations are not always sufficient to explain the decisions of deep neural networks e.g., in Figure \ref{fig:results-1}, single level explanation only provides reasoning behind the prediction \textit{``playing daf''}. However, someone who is not familiar with a \textit{daf} may not understand the explanation. Instead, when we provide explanations at more abstract levels i.e., parent and grandparent level. By looking at the explanations for parent class one can infer that it involves \textit{``percussion''} and from the grandparent that it is some sort of musical \textit{``instrument''}. We propose to incorporate this powerful mechanism for providing explanations based on hierarchical prior knowledge into networks for video action recognition.

\subsection{Hierarchical Prototype Explainer}
Figure \ref{fig:Architecture} gives an overview of our proposed Hierarchical prototype explainer (HIPE) for video action recognition. HIPE consists of a 3D-CNN backbone $f$ for extracting features from the video frames, and a hierarchical prototype layer $g_p$ for learning prototypes for each frame. The prototype layer is followed by a fully connected layer $h$ that combines the prototype similarity scores and maps them to the shared hyperbolic space through exponential mapping. Prior knowledge about the relations between actions, in the form of the action hierarchy, are projected to the shared space through discriminative embeddings. Subsequently, we use hyperbolic learning to obtain hierarchical prototypes that enable multi-level explainability.

As the backbone architecture, we use the video action classification network 3D-Resnet \cite{hara3dcnns}. For each input video $v \in \mathbb{R}^{W\times H \times T \times 3}$ with $T$ frames it extracts video features $Z\in \mathbb{R}^{W_0\times H_0 \times T_0 \times D}$ with the spatial resolution $W_0\times H_0$, frames $T_0$ and channels $D$. A key aspect of this backbone is that $T_0 < T$ due to temporal pooling, as such the features $Z$ are extracted for segments rather than individual frames. Because of the temporal pooling, the prototypes learned by HIPE are spatio-temporal thereby explaining which parts of the segment are indicative of the action in the video.

\subsection{Hierarchical Prototype Layer} 
 Given the features extracted from the 3D-Resnet $Z$, two layers of $1\times1\times1$ convolutions with the LeakyReLU activations  are added for adjusting the number of channels for the top layer. Hierarchical prototypes are learned by incorporating the prior hierarchical knowledge about actions into the network by representing them as embeddings. We optimize hierarchical prototypes by aligning them to the action embeddings in hyperbolic space. 

Given the set of action classes $\mathcal{A}=\{1,2,...,|\mathcal{A}|\}$, in hierarchical action recognition we also consider their ancestor classes $\mathcal{H}=\{|\mathcal{A}|+1,|\mathcal{A}|+2,...,|\mathcal{A}|+|\mathcal{H}|\}$, which allows us to construct a hierarchical tree with three levels, i.e., grandparent, parent, and child (see Figure \ref{fig:Architecture} right). These hierarchies can be easily defined by considering relations between classes, and do not require annotation of individual instances.  The process of embedding the hierarchies is performed once, offline, per dataset. However, this process can easily be repeated for alternative hierarchies. See supplementary material for further details.

For each child $\mathcal{A}$ and its parent $\mathcal{H}$ action, the network learns $m$ and $n$ prototypes respectively $P=\{p_j\}_{j=1}^{m+n}$, whose shape is ${W_1\times H_1 \times T_1 \times D}$ with $W_1 \leq W_0$, $H_1 \leq H_0$ and $T_1 \leq T_0$. As such each prototype represents a spatio-temporal part of the video. Given the convolutional output $Z=f(v)$ and prototypes $p$, a prototype layer $g_p$ computes the distances between each prototype $p_j$ and the patches from $Z$ and converts them to the similarity scores using
\begin{equation}
\label{eq3.0.1}
    g_p(p_j , Z) = \max_{z \in Z}\log\frac{(||z-p_j||_2^2 +1)}{(||z-p_j||_2^2 +\epsilon)} , \epsilon>0
\end{equation}
The distances between each prototype and the patch determine the extent to which a prototype is present in the input. We expect to learn different prototypes for child, parent and its grandparent e.g. in Figure \ref{fig:Architecture} the prototype learned for the \textit{applying eye make} child class is an eye, for the parent it is focusing on hairs because of the parent class \textit{self grooming} and shows hand for the grandparent \textit{human object interaction} class.
We then multiply similarity scores with the weights of a fully connected layer $h$ to obtain embeddings to be projected in the hyperbolic joint space for learning hierarchical prototypes.

\subsection{Hierarchical Video Embeddings} The embeddings $\vh=h(g_p(p,f(v)))$ obtained from the fully connected layer are in the Euclidean space and can not be directly mapped into the hyperbolic embedding space, therefore, we use exponential mapping \cite{ganea2018hyperbolic} to map video embeddings into the hyperbolic space.  
\begin{equation}
\label{eq:3.3.1}
    \exp_\vx(\vh)= \vx \oplus \Bigl( \tanh \Bigl(\frac{||\vh||}{1-\|\vx\|^2}\Bigr)\frac{\vh}{||\vh||}\Bigr) 
\end{equation}
where $\oplus$ indicates the $1-$curved Mobius addition, $\vx$ is the tangent point connecting tangent space $\mathcal{T}_{\vzero}\mathbb{D}^n$ to $\mathbb{D}^n$. Different values of $x$ lead to different tangent spaces, to avoid any ambiguities we set $\vx=\vzero$ and project the video embeddings to the hyperbolic space for matching with the hierarchical actions.

\subsection{Training}
Our training process consists of a multi-step procedure: In the initial epochs we perform warm-up of the newly added layers. Following the warm-up, we train the entire network end-to-end. Every $10$ epochs we update the prototype layer only, followed by a phase of fine-tuning the layers after the prototype layer. 

\myparagraph{Video and Action Matching in the Hyperbolic Space}

We aim to learn a latent space where patches important for classification are clustered around similar prototypes.  In order to learn hierarchical prototypes we optimize the prototypes $P=\{p_j\}_{j=1}^{m+n}$ to match videos to hyperbolic action embeddings, hence our optimization is supervised by $\vPhi \in \DD^{n\times(|\CA|)}$. Let $\{(v_i,y_i)\}_{i=1}^N$ be the training set, where $v \in \mathbb{R}^{W\times H \times T \times 3}$ and $y_i \in \mathcal{A}$. 
Our goal is to solve:
\begin{equation}
\label{eq:3.4.1}
    \mathcal{L}_{crs} + \lambda_1 \mathcal{L}_{cls} + \lambda_2 \mathcal{L}_{sep}
\end{equation}
 \myparagraph{Hierarchical Cross Entropy.} The first term in our loss is the hierarchical cross-entropy loss $\mathcal{L}_{crs}$ which penalizes the misclassification, and is defined as:
\begin{equation}
\label{eq:3.4.2}
    \mathcal{L}_{crs} = \frac{1}{N}\sum_{i=1}^N\sum_{k=1}^K y_{ik}\log p(y=k|v)
\end{equation}
The Softmax in the cross entropy is defined as the negative distance between video embeddings and the hierarchical action embeddings in the hyperbolic space:
\begin{equation}
\label{eq:3.4.3}
    p(y=k|v) = \frac{\exp(-d(\vh_e),\mathbf{\Phi}_k)}{\sum_{k'}\exp(-d(\vh_e),\mathbf{\Phi}_k))},
\end{equation}
where $\vh_e=\exp_{\vzero}(\vh)$ is applying exponential map to the fully connected layer output $\vh$. 

 \myparagraph{Hierarchical Clustering.}
 In order to provide meaningful explanations at each level of hierarchy our hierarchical clustering cost encourages input frames to have at least one patch from features to be closer to a child, parent or grandparent class prototype.
 \begin{equation}
     \label{eq:3.4.4}
      \mathcal{L}_{cls} = \frac{1}{N}\sum_{i=1}^N \min_{j:p_j\in P_{|\mathcal{A}|+|\mathcal{H}|}} \min_{z\in patches(f(v_i))}||z-p_j||_2^2 
 \end{equation}
 \myparagraph{Hierarchical Separation.}
 Our hierarchical separation cost encourages the latent patches of the frames to stay away from the prototypes not belonging to the same child class or parent class or grandparent class.
\begin{equation}
    \label{eq:3.4.3}
      \mathcal{L}_{sep} = -\frac{1}{N}\sum_{i=1}^N \min_{j:p_j\notin P_{|\mathcal{A}|+|\mathcal{H}|}} \min_{z\in patches(f(v_i))}||z-p_j||_2^2 
\end{equation}

\begin{figure*}
    \centering
    \includegraphics[width=0.9\linewidth]{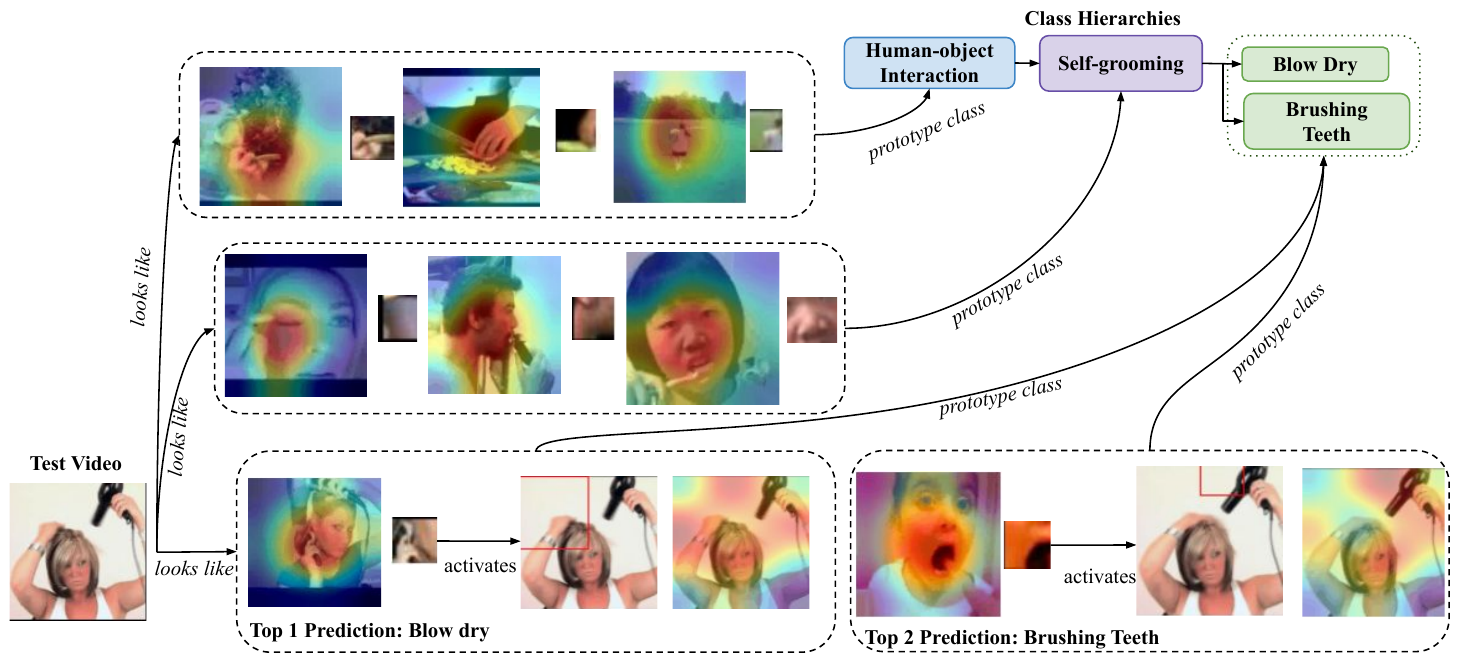}
    \caption{\textbf{Hierarchical Explanations.} This example shows the prototypes from grandparent \textit{human-object interaction} class, parent \textit{self-grooming} class and ground truth \textit{blow dry} class, we also observe that the top 2 prediction for the model is its sibling \textit{brushing teeth} class. This conforms that our model is learning hierarchical relations between classes. } 
    \label{fig:results-3}
    
\end{figure*}
\subsection{Updating Hierarchical Prototype Layer}
We project prototypes onto the closest video features from the training videos. We do so for child, parent, and grandparent action categories. Mathematically, for the prototypes $p_j$ from child, parent and grandparent class i.e. $p_j \in P_{|\mathcal{A}|+|\mathcal{H}|}$, we update the prototype layer as:

\begin{equation}
\label{eq:3.4.4}
    p_j \leftarrow \underset{z\in \mathcal{Z}_j}{\text{argmin}}||z-p_j||_2
\end{equation}

where $\mathcal{Z}_j= \{\tilde{z}: \tilde{z} \in patches(f(v_i)) \, \forall i \, \text{s.t.} \, y_i = |\mathcal{A}|+|\mathcal{H}| \}$. Our prototype layer is updated not only with the prototypes belonging to the child class but also with the parent and (grand)parent classes enabling the learning of hierarchical relations between classes.

\subsection{Hierarchical Prototype Visualization}
To construct the visualizations the learned prototypes are mapped to the spatio-temporal input space. We select the patch which highly activates for the prototype $p_j$ by forwarding the input $v$ through the network and upsampling the activation map generated by the prototype layer $g_p(p_j, Z)$ both spatially and temporally (for videos). We visualize $p_j$ for child, parent, and grandparent classes providing explanations at all levels of the hierarchy.

\section{Experimental Setup}

\subsection{Datasets}
To evaluate HIPE for videos we conduct experiments on two video datasets: UCF-101 \cite{soomro2012ucf101} and Activity-Net1.3 \cite{caba2015activitynet}.

\myparagraph{Hierarchical UCF-101.} UCF-101 \cite{soomro2012ucf101} contains 13,320 videos belonging to 101 action categories with a total length of 27 hours. We define two additional levels of hierarchy with the number of classes at level one, two, and three being 5, 20, and 101 respectively. The classes at the third level of the hierarchy are the 101 original classes of the dataset. The full hierarchy is included in the supplementary material.

\myparagraph{Hierarchical ActivityNet.}
ActivityNet \cite{caba2015activitynet} contains 14,950 untrimmed videos with each video consisting of one or more action segments belonging to 200 action classes with a total length of approximately 648 hours. We use 10,024 videos for training and 4,926 for validation. We follow \cite{hara3dcnns} and train and test our model on trimmed videos to determine video-level accuracy. We follow \cite{long2020searching} to define the class-level action hierarchies using the hierarchies that come with the dataset. It contains 200, 38, and 6 classes in level one, two, and three respectively.

\subsection{Implementation Details}
The hierarchical action embeddings are generated by training the model with Reimannian Adam optimizer \cite{becigneul2018riemannian}, implemented with \textit{geoopt} and Pytorch \cite{paszke2019pytorch}. Apart from the one-time offline step of generating the hierarchical action embedding HIPE is trained in an end-to-end fashion. For feature extraction, we used Resnet-3D-18 \cite{hara3dcnns} pre-trained on Kinetics \cite{carreira2017quo} and added two $1\times 1$ convolutional layers with the LeakyReLU, a prototype layer and the final embedding layer. We perform prototype projection and visualization every 10 epochs. 

We report results on two variations of HIPE:  we compare between prototype projection with 5 prototypes per class and 10 prototypes per class, to explore whether this additional supervision makes it possible to use fewer prototypes. For comparison, we adapt ProtoPNet \cite{chen2019looks} to videos by replacing the 2D ResNet backbone with a 3D ResNet.

\myparagraph{Evaluation Metrics.}
We report the performance at both clip level and video level. The clip level accuracy is the rate of correct prediction of each clip, while the video level accuracy is the majority vote of the predictions over all the clips within the video.
Additionally, to show the benefit of using hierarchical learning, we report accuracy for three metrics: the class accuracy is calculated as the rate of correct prediction in the hierarchical space 0-hop away from the ground-truth, the sibling accuracy as the rate of correct prediction 2-hops away from the ground-truth, and the cousin accuracy as the 4-hops correct prediction rate. Higher performance on the sibling and cousin metrics indicates that misclassifications are to hierarchically nearby, and therefore, semantically similar classes.

% \section{Experiments}
\section{Visual Explanations} 
\vspace{-2mm}

\begin{figure*}
    \centering
    \includegraphics[width=0.9\linewidth]{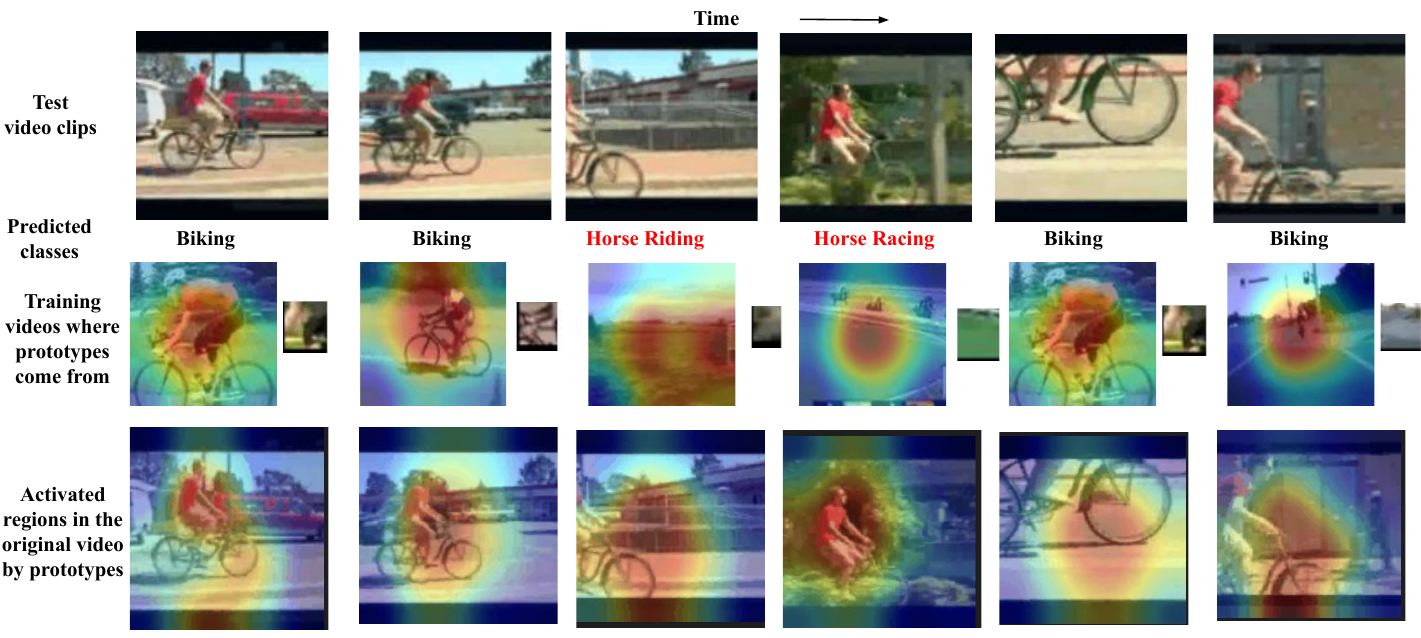}
    \caption{\textbf{Spatio-temporal Explanations.} Explanations for a video with six clips. Top row: one frame per clip. Second row: class predictions. Bottom row: training video frames where prototypes came from and classification is based on. When the network is focusing on more abstract concepts like grass and barrier bars it gets misclassified into \textit{horse riding} class. Similarly, when it is focusing on grass in the field it gets misclassified into \textit{horse racing} class. While when it focuses on more concrete concepts like bicycle  tyres or handle it gets correctly classified. }
    
    \label{fig:results-2}
    
\end{figure*}

\begin{figure}
    \centering
    \includegraphics[width=\linewidth]{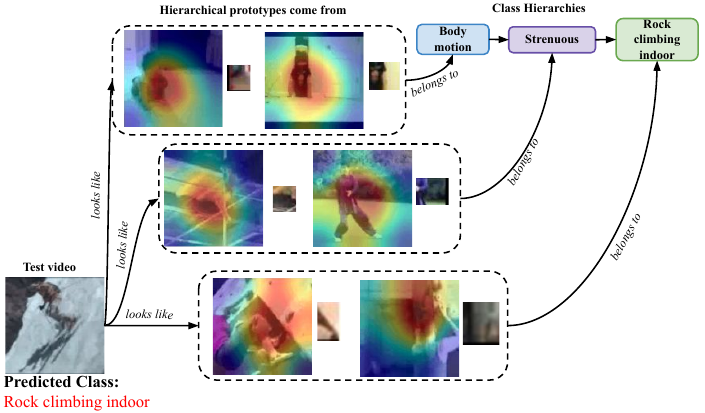}
    \caption{\textbf{Effectiveness in case of failure.} Our multi-level explanations provide useful information even in the case of misclassification through the prototypes learned for parent and grandparent classes.}
    \label{fig:results-4}
    \vspace{-2mm}
\end{figure}
\begin{table*}
\begin{center}
\begin{tabular}{l  c c  c c c }
\hline
& Network & {Accuracy} & {Sibling Accuracy} & {Cousin Accuracy} & \shortstack{\# of prototypes\\ per class} \\
\hline 
 \multirow{2}{*}{\shortstack{Non-Interpretable \\ Models}} & 3D-Resnet  \cite{hara3dcnns} & 83.34 & 89.73 & 93.62 & -\\
% & Resnet-Hyperbolic   &  &  &  & -\\
& Resnet-Hyperbolic \cite{long2020searching}  & 82.64 & 89.99 & 93.28 & -\\
\hline
\multirow{3}{*}{\shortstack{Interpretable \\ Models}}  & ProtoPNet  \cite{chen2019looks}  & 78.30 &85.92 & 90.98 & 10\\
% & Hierarchical ProtoPNet   & 79.49  & 88.60 & 92.79 & 10 \\
& \shortstack{HIPE } & 79.45 & 88.88 & 92.73 & 5\\   
& \shortstack{HIPE } & \textbf{80.40} & \textbf{89.30} & \textbf{93.02} & 10\\    
\hline
\end{tabular}
\end{center}
\caption{\textbf{Clip level accuracy comparison for different models on UCF-101 videos.} We observe that HIPE with 10 prototypes per class recovers the drop due to accuracy-explainability trade off significantly while providing multi-level explanations.}\label{table:Acc-UCF}
\vspace{-2mm}
\end{table*} 
\label{subsec:visual_exp}

\myparagraph{Hierarchical Explanations.} Figure \ref{fig:results-3} shows an example of multi-level explanations provided by HIPE. Our model learns to represent the video clip features as hierarchical prototypes that belong to grandparent, parent and child classes. For example, in Figure \ref{fig:results-3} our model has learned prototypes (only three out of ten prototypes shown for better presentation) from the grandparent class \textit{human-object interaction},  parent class \textit{self-grooming}, and the action class \textit{blow dry} (only one prototype and its activation on the original video shown). We also observe that the second most likely prediction is its sibling class \textit{brushing teeth}.

\myparagraph{Spatio-temporal Explanations.} Figure \ref{fig:results-2} shows explanations for a video with six clips each clip with 16 frames. The top row shows single frames from
different clips of a video. In the second row we observe that the model fails to predict correct classes for third and fourth clip. The bottom row shows training video frames where prototypes came from (one prototype per clip shown for better visibility) and classification is based on. We see that when the network is focusing on the more concrete concepts like the tyres of the bicycle, or handle it gets correctly classified. While when the network is focusing on the grass and barrier bars, and it activates bars in the background of the cyclist in original clip, it gets misclassified into the \textit{horse riding class}. Similarly, when it is focusing on grass in the field, it activates greenery in the background of the cyclist in original clip, it gets misclassified into \textit{horse racing class}. The parent and grandparent classes for all the clips are still the same i.e., \textit{riding sports} and \textit{sports} respectively, giving us useful information even in case of misclassification.

\myparagraph{Effectiveness of Hierarchical Explanations in case of Failure.} In Figure \ref{fig:results-4} we show another scenario where the multi-level explanations are useful. We see that the original skiing video is misclassified into the \textit{rock climbing indoor} class. However, for the more abstract explanations we can observe that its parent class \textit{strenuous sports} and grandparent class \textit{body motion} are correctly recognized. Hence our hierarchical explanations give us useful information even in the case of misclassification.

\section{Accuracy-Explainability Trade Off}
In order to show that HIPE reduces accuracy-explainability trade off here we present quantitative comparison of our model with the interpretable and non-interpretable baselines.

\begin{table*}
\begin{center}
\begin{tabular}{l  c c  c c c }
\hline
& Network & {Accuracy} & {Sibling Accuracy} & {Cousin Accuracy}& \shortstack{\# of prototypes\\ per class} \\
\hline 
 \multirow{2}{*}{\shortstack{Non-Interpretable \\ Models}} & 3D-Resnet  \cite{hara3dcnns} & 49.99 & 51.59 & 63.88 & -\\
& Hyperbolic-Resnet \cite{long2020searching}  & 49.95 & 52.03 & 65.44 & - \\
\hline
\multirow{3}{*}{\shortstack{Interpretable \\ Models}}  & ProtoPNet  \cite{chen2019looks}   & 46.06 & 47.74& 61.67 & 10 \\
% & Hierarchical ProtoPNet    & 46.02  & 48.76 & 62.77 & 10 \\
& \shortstack{HIPE } & 45.67 & 48.72 & 62.84 & 5\\  
& \shortstack{HIPE } & \textbf{46.26}  & \textbf{48.93} & \textbf{62.97} & 10\\ 
\hline
\end{tabular}
\end{center}
\caption{\textbf{Clip level accuracy comparison for different models on ActivityNet videos.} We observe that HIPE with 10 prototypes per class recovers the drop for siblings and cousins and shows comparable performance with regular ProtoPNet for class accuracy while providing multi-level explanations.}\label{table:Acc-ActivityNet}
\end{table*}

\begin{table*}
\begin{center}
\begin{tabular}{l  c c c c c c }
\hline
& Network & {\shortstack{ UCF-101\\ Accuracy}} & {\shortstack{ActivityNet\\ Accuracy}} & \shortstack{\# of prototypes\\ per class}\\
\hline 
 \multirow{2}{*}{\shortstack{Non-Interpretable \\ Models}} & 3D-Resnet  \cite{hara3dcnns} & 87.92 & 69.15 & -\\
&Resnet-Hyperbolic \cite{long2020searching}  & 87.15 & 70.19 & -\\
\hline
\multirow{4}{*}{\shortstack{Interpretable \\ Models}} &ProtoPNet  \cite{chen2019looks}  & 84.48 & 66.46 & 10\\
% &Hierarchical ProtoPNet   & 84.03 & 64.66& 10 \\
% &Hierarchical ProtoPNet (w/o push update)   & 85.20 &  &10\\
&HIPE & 84.87 & 63.82 & 5\\
&HIPE & \textbf{86.22}  & \textbf{66.48} & 10\\
\hline
\end{tabular}
\end{center}
\caption{\textbf{Video level accuracy comparison for different models on UCF-101 and ActivityNet videos.} We observe that HIPE with 10 prototypes per class recovers the drop at video level significantly for UCF-101 and shows comparable performance with the regular ProtoPNet for ActivityNet while providing multi-level explanations.}\label{table:Acc-Videolevel}
\vspace{-3mm}
\end{table*} 
\myparagraph{Non-Interpretable Models.} The performance of non-interpretable models on UCF-101 and ActivityNet are shown in the top two rows of Tables \ref{table:Acc-UCF}, \ref{table:Acc-ActivityNet}, and \ref{table:Acc-Videolevel}. For fair comparison both non-interpretable models, a regular Resnet \cite{hara3dcnns} and a hyperbolic Resnet \cite{long2020searching}, are trained end-to-end with the same data augmentations and an equal number of epochs. The only difference between the two non-interpretable models is that for the Resnet model the categories are separated through euclidean hyperplanes while the Resnet-Hyperbolic utilizes hyperbolic embedding space to separate categories. Our results for UCF-101 show that both a regular Resnet and the hyperbolic Resnet perform similarly at clip level (Table \ref{table:Acc-UCF}) and video level (Table \ref{table:Acc-Videolevel}).

For ActivityNet the clip-level class accuracy (Table \ref{table:Acc-ActivityNet}) is comparable across both networks, however, due to the hierarchical learning of hyperbolic Resnet it shows better sibling and cousin accuracy, additionally, it shows an improvement for class accuracy at the video level (see Table \ref{table:Acc-Videolevel}). Overall, we see comparable performance for the non-interpretable networks on UCF-101 and improvements for the Hyperbolic networks on ActivityNet.

\myparagraph{Interpretable Models.} The performance of interpretable models on UCF-101 and ActivityNet are shown in the bottom three rows of Table \ref{table:Acc-UCF}, \ref{table:Acc-ActivityNet}, and \ref{table:Acc-Videolevel}. We report the results for a regular ProtoPNet \cite{chen2019looks} adapted for videos and the variations of HIPE.

For UCF-101, with a regular ProtoPNet with 10 prototypes per class, the accuracy drops considerably: the clip-level class accuracy drops to $78.30$ and the video-level accuracy to $84.48$. This is because of the explainability-accuracy trade off common in explainable-AI, also reported in \cite{chen2019looks}. In contrast, HIPE with 5 prototypes per class is much less affected and recovers the drop by $1.15$ for class accuracy, $2.96$, and $1.75$ for sibling and cousin accuracies respectively.  Increasing the number of prototypes to 10 per class further improves the performance by $2.10$, $3.38$, and $2.04$ for class, sibling, and cousin accuracies respectively. Moreover, the performance at the video level reaches $86.22$ (see Table \ref{table:Acc-Videolevel}). Hence, both variations of our proposed HIPE reduce the accuracy drop. 

On ActivityNet (see Table \ref{table:Acc-ActivityNet}) we observe a clear accuracy-explainability trade off for the regular ProtoPNet, with drops in both the clip and video level accuracies. However, whilst HIPE shows a similar drop in class accuracy we can observe that it partially recovers from this drop on the sibling and cousin metrics. This behavior holds for both the HIPE with 5 prototypes, and for the variant with 10 prototypes per class, we even see improvements for the sibling and cousin metrics of $1.19$ and $1.3$ respectively. Whilst ActivityNet remains challenging, an improvement in sibling and cousin accuracies is directly beneficial to the explainability as demonstrated in Section~\ref{subsec:visual_exp}.

Hence, we can observe that on both datasets HIPE is less affected by the accuracy-explainability trade off whilst also providing multi-level explanations.

%\begin{figure*}
%    \centering
%    \includegraphics[width=\linewidth]{images/Results_HProtoPNet_2.pdf}
%    \caption{Results \textcolor{red}{remove this one from the paper??}}
%    \label{fig:results-2}
%\end{figure*}

\section{Conclusion}

In this work, we proposed Hierarchical prototype explainer for video action recognition. By learning hierarchical prototypes we are able to provide explanations at multiple levels of granularity, not only explaining why it is classified as a certain class, but also what spatiotemporal parts contribute to it belonging to parent categories. Our results show that HIPE outperforms a prior non-hierarchical approach on UCF-101, whilst performing equally well on ActivityNet. Additionally, we demonstrate our multi-level explanations that make it possible to see which spatiotemporal parts contribute to grandparent, parent, and class-level classifications. Our hierarchical approach thereby provides richer explanations whilst compromising less performance to gain explainability.

{\small
\bibliographystyle{ieee_fullname}
\bibliography{egbib}

\begin{thebibliography}{10}\itemsep=-1pt

\bibitem{atigh2022hyperbolic}
Mina~Ghadimi Atigh, Julian Schoep, Erman Acar, Nanne van Noord, and Pascal
  Mettes.
\newblock Hyperbolic image segmentation.
\newblock In {\em Proceedings of the IEEE/CVF Conference on Computer Vision and
  Pattern Recognition}, pages 4453--4462, 2022.

\bibitem{bargal2018excitation}
Sarah~Adel Bargal, Andrea Zunino, Donghyun Kim, Jianming Zhang, Vittorio
  Murino, and Stan Sclaroff.
\newblock Excitation backprop for rnns.
\newblock In {\em Proceedings of the IEEE Conference on Computer Vision and
  Pattern Recognition}, pages 1440--1449, 2018.

\bibitem{becigneul2018riemannian}
Gary B{\'e}cigneul and Octavian-Eugen Ganea.
\newblock Riemannian adaptive optimization methods.
\newblock {\em arXiv preprint arXiv:1810.00760}, 2018.

\bibitem{bertinetto2016fully}
Luca Bertinetto, Jack Valmadre, Joao~F Henriques, Andrea Vedaldi, and Philip~HS
  Torr.
\newblock Fully-convolutional siamese networks for object tracking.
\newblock In {\em European conference on computer vision}, pages 850--865.
  Springer, 2016.

\bibitem{caba2015activitynet}
Fabian Caba~Heilbron, Victor Escorcia, Bernard Ghanem, and Juan Carlos~Niebles.
\newblock Activitynet: A large-scale video benchmark for human activity
  understanding.
\newblock In {\em Proceedings of the ieee conference on computer vision and
  pattern recognition}, pages 961--970, 2015.

\bibitem{carreira2017quo}
Joao Carreira and Andrew Zisserman.
\newblock Quo vadis, action recognition? a new model and the kinetics dataset.
\newblock In {\em proceedings of the IEEE Conference on Computer Vision and
  Pattern Recognition}, pages 6299--6308, 2017.

\bibitem{chami2019hyperbolic}
Ines Chami, Zhitao Ying, Christopher R{\'e}, and Jure Leskovec.
\newblock Hyperbolic graph convolutional neural networks.
\newblock {\em Advances in neural information processing systems}, 32, 2019.

\bibitem{chen2019looks}
Chaofan Chen, Oscar Li, Daniel Tao, Alina Barnett, Cynthia Rudin, and
  Jonathan~K Su.
\newblock This looks like that: deep learning for interpretable image
  recognition.
\newblock {\em Advances in neural information processing systems}, 32, 2019.

\bibitem{dhingra2018embedding}
Bhuwan Dhingra, Christopher~J Shallue, Mohammad Norouzi, Andrew~M Dai, and
  George~E Dahl.
\newblock Embedding text in hyperbolic spaces.
\newblock {\em arXiv preprint arXiv:1806.04313}, 2018.

\bibitem{donnelly2022deformable}
Jon Donnelly, Alina~Jade Barnett, and Chaofan Chen.
\newblock Deformable protopnet: An interpretable image classifier using
  deformable prototypes.
\newblock In {\em Proceedings of the IEEE/CVF Conference on Computer Vision and
  Pattern Recognition}, pages 10265--10275, 2022.

\bibitem{fang2021kernel}
Pengfei Fang, Mehrtash Harandi, and Lars Petersson.
\newblock Kernel methods in hyperbolic spaces.
\newblock In {\em Proceedings of the IEEE/CVF International Conference on
  Computer Vision}, pages 10665--10674, 2021.

\bibitem{ganea2018hyperbolic}
Octavian Ganea, Gary B{\'e}cigneul, and Thomas Hofmann.
\newblock Hyperbolic entailment cones for learning hierarchical embeddings.
\newblock In {\em International Conference on Machine Learning}, pages
  1646--1655. PMLR, 2018.

\bibitem{ghadimi2021hyperbolic}
Mina Ghadimi~Atigh, Martin Keller-Ressel, and Pascal Mettes.
\newblock Hyperbolic busemann learning with ideal prototypes.
\newblock {\em Advances in Neural Information Processing Systems}, 34:103--115,
  2021.

\bibitem{10.1145/3372278.3390672}
Sadaf Gulshad and Arnold Smeulders.
\newblock Explaining with counter visual attributes and examples.
\newblock In {\em Proceedings of the 2020 International Conference on
  Multimedia Retrieval}, ICMR '20, 2020.

\bibitem{hara3dcnns}
Kensho Hara, Hirokatsu Kataoka, and Yutaka Satoh.
\newblock Can spatiotemporal 3d cnns retrace the history of 2d cnns and
  imagenet?
\newblock In {\em Proceedings of the IEEE Conference on Computer Vision and
  Pattern Recognition (CVPR)}, pages 6546--6555, 2018.

\bibitem{Hendricks_2018_ECCV}
Lisa~Anne Hendricks, Ronghang Hu, Trevor Darrell, and Zeynep Akata.
\newblock Grounding visual explanations.
\newblock In {\em ECCV}, September 2018.

\bibitem{karpathy2015visualizing}
Andrej Karpathy, Justin Johnson, and Li Fei-Fei.
\newblock Visualizing and understanding recurrent networks.
\newblock {\em arXiv preprint arXiv:1506.02078}, 2015.

\bibitem{KarpathyCVPR14}
Andrej Karpathy, George Toderici, Sanketh Shetty, Thomas Leung, Rahul
  Sukthankar, and Li Fei-Fei.
\newblock Large-scale video classification with convolutional neural networks.
\newblock In {\em CVPR}, 2014.

\bibitem{khrulkov2020hyperbolic}
Valentin Khrulkov, Leyla Mirvakhabova, Evgeniya Ustinova, Ivan Oseledets, and
  Victor Lempitsky.
\newblock Hyperbolic image embeddings.
\newblock In {\em Proceedings of the IEEE/CVF Conference on Computer Vision and
  Pattern Recognition}, pages 6418--6428, 2020.

\bibitem{kim2018interpretability}
Been Kim, Martin Wattenberg, Justin Gilmer, Carrie Cai, James Wexler, Fernanda
  Viegas, et~al.
\newblock Interpretability beyond feature attribution: Quantitative testing
  with concept activation vectors (tcav).
\newblock In {\em International conference on machine learning}, pages
  2668--2677. PMLR, 2018.

\bibitem{Kim_2018_ECCV}
Jinkyu Kim, Anna Rohrbach, Trevor Darrell, John Canny, and Zeynep Akata.
\newblock Textual explanations for self-driving vehicles.
\newblock In {\em ECCV}, September 2018.

\bibitem{koh2020concept}
Pang~Wei Koh, Thao Nguyen, Yew~Siang Tang, Stephen Mussmann, Emma Pierson, Been
  Kim, and Percy Liang.
\newblock Concept bottleneck models.
\newblock In {\em International Conference on Machine Learning}, pages
  5338--5348. PMLR, 2020.

\bibitem{KrauseStarkDengFei-Fei_3DRR2013}
Jonathan Krause, Michael Stark, Jia Deng, and Li Fei-Fei.
\newblock 3d object representations for fine-grained categorization.
\newblock In {\em 4th International IEEE Workshop on 3D Representation and
  Recognition (3dRR-13)}, Sydney, Australia, 2013.

\bibitem{li2018deep}
Oscar Li, Hao Liu, Chaofan Chen, and Cynthia Rudin.
\newblock Deep learning for case-based reasoning through prototypes: A neural
  network that explains its predictions.
\newblock In {\em Proceedings of the AAAI Conference on Artificial
  Intelligence}, volume~32, 2018.

\bibitem{li2021towards}
Zhenqiang Li, Weimin Wang, Zuoyue Li, Yifei Huang, and Yoichi Sato.
\newblock Towards visually explaining video understanding networks with
  perturbation.
\newblock In {\em Proceedings of the IEEE/CVF Winter Conference on Applications
  of Computer Vision}, pages 1120--1129, 2021.

\bibitem{lin2021gradient}
Xinmiao Lin, Wentao Bao, Matthew Wright, and Yu Kong.
\newblock Gradient frequency modulation for visually explaining video
  understanding models.
\newblock {\em arXiv preprint arXiv:2111.01215}, 2021.

\bibitem{liu2020hyperbolic}
Shaoteng Liu, Jingjing Chen, Liangming Pan, Chong-Wah Ngo, Tat-Seng Chua, and
  Yu-Gang Jiang.
\newblock Hyperbolic visual embedding learning for zero-shot recognition.
\newblock In {\em Proceedings of the IEEE/CVF conference on computer vision and
  pattern recognition}, pages 9273--9281, 2020.

\bibitem{long2020searching}
Teng Long, Pascal Mettes, Heng~Tao Shen, and Cees~GM Snoek.
\newblock Searching for actions on the hyperbole.
\newblock In {\em Proceedings of the IEEE/CVF Conference on Computer Vision and
  Pattern Recognition}, pages 1141--1150, 2020.

\bibitem{losch2019interpretability}
Max Losch, Mario Fritz, and Bernt Schiele.
\newblock Interpretability beyond classification output: Semantic bottleneck
  networks.
\newblock {\em arXiv preprint arXiv:1907.10882}, 2019.

\bibitem{lundberg2017unified}
Scott~M Lundberg and Su-In Lee.
\newblock A unified approach to interpreting model predictions.
\newblock {\em Advances in neural information processing systems}, 30, 2017.

\bibitem{mcclelland2003parallel}
James~L McClelland and Timothy~T Rogers.
\newblock The parallel distributed processing approach to semantic cognition.
\newblock {\em Nature reviews neuroscience}, 4(4):310--322, 2003.

\bibitem{minsky1982semantic}
Marvin Minsky.
\newblock Semantic information processing.
\newblock 1982.

\bibitem{nauta2021neural}
Meike Nauta, Ron van Bree, and Christin Seifert.
\newblock Neural prototype trees for interpretable fine-grained image
  recognition.
\newblock In {\em Proceedings of the IEEE/CVF Conference on Computer Vision and
  Pattern Recognition}, pages 14933--14943, 2021.

\bibitem{nickel2017poincare}
Maximillian Nickel and Douwe Kiela.
\newblock Poincar{\'e} embeddings for learning hierarchical representations.
\newblock {\em Advances in neural information processing systems}, 30, 2017.

\bibitem{Nickel:2017wz}
Maximilian Nickel and Douwe Kiela.
\newblock {Poincar{\'e} Embeddings for Learning Hierarchical Representations.}
\newblock In {\em NeurIPS}, 2017.

\bibitem{oikarinenlabel}
Tuomas Oikarinen, Subhro Das, Lam~M Nguyen, and Tsui-Wei Weng.
\newblock Label-free concept bottleneck models.
\newblock In {\em International Conference on Learning Representations}.

\bibitem{paszke2019pytorch}
Adam Paszke, Sam Gross, Francisco Massa, Adam Lerer, James Bradbury, Gregory
  Chanan, Trevor Killeen, Zeming Lin, Natalia Gimelshein, Luca Antiga, et~al.
\newblock Pytorch: An imperative style, high-performance deep learning library.
\newblock {\em Advances in neural information processing systems}, 32, 2019.

\bibitem{peng2016multi}
Xiaojiang Peng and Cordelia Schmid.
\newblock Multi-region two-stream r-cnn for action detection.
\newblock In {\em European conference on computer vision}, pages 744--759.
  Springer, 2016.

\bibitem{ribeiro2016should}
Marco~Tulio Ribeiro, Sameer Singh, and Carlos Guestrin.
\newblock " why should i trust you?" explaining the predictions of any
  classifier.
\newblock In {\em Proceedings of the 22nd ACM SIGKDD international conference
  on knowledge discovery and data mining}, pages 1135--1144, 2016.

\bibitem{rymarczyk2020protopshare}
Dawid Rymarczyk, {\L}ukasz Struski, Jacek Tabor, and Bartosz Zieli{\'n}ski.
\newblock Protopshare: Prototype sharing for interpretable image classification
  and similarity discovery.
\newblock {\em arXiv preprint arXiv:2011.14340}, 2020.

\bibitem{simonyan2014two}
Karen Simonyan and Andrew Zisserman.
\newblock Two-stream convolutional networks for action recognition in videos.
\newblock {\em Advances in neural information processing systems}, 27, 2014.

\bibitem{singh2017online}
Gurkirt Singh, Suman Saha, Michael Sapienza, Philip~HS Torr, and Fabio
  Cuzzolin.
\newblock Online real-time multiple spatiotemporal action localisation and
  prediction.
\newblock In {\em Proceedings of the IEEE International Conference on Computer
  Vision}, pages 3637--3646, 2017.

\bibitem{soomro2012ucf101}
Khurram Soomro, Amir~Roshan Zamir, and Mubarak Shah.
\newblock Ucf101: A dataset of 101 human actions classes from videos in the
  wild.
\newblock {\em arXiv preprint arXiv:1212.0402}, 2012.

\bibitem{stergiou2019class}
Alexandros Stergiou, Georgios Kapidis, Grigorios Kalliatakis, Christos
  Chrysoulas, Ronald Poppe, and Remco Veltkamp.
\newblock Class feature pyramids for video explanation.
\newblock In {\em 2019 IEEE/CVF International Conference on Computer Vision
  Workshop (ICCVW)}, pages 4255--4264. IEEE, 2019.

\bibitem{stergiou2019saliency}
Alexandros Stergiou, Georgios Kapidis, Grigorios Kalliatakis, Christos
  Chrysoulas, Remco Veltkamp, and Ronald Poppe.
\newblock Saliency tubes: Visual explanations for spatio-temporal convolutions.
\newblock In {\em 2019 IEEE International Conference on Image Processing
  (ICIP)}, pages 1830--1834. IEEE, 2019.

\bibitem{suris2021learning}
D{\'\i}dac Sur{\'\i}s, Ruoshi Liu, and Carl Vondrick.
\newblock Learning the predictability of the future.
\newblock In {\em Proceedings of the IEEE/CVF Conference on Computer Vision and
  Pattern Recognition}, pages 12607--12617, 2021.

\bibitem{tifrea2018poincar}
Alexandru Tifrea, Gary B{\'e}cigneul, and Octavian-Eugen Ganea.
\newblock Poincar$\backslash$'e glove: Hyperbolic word embeddings.
\newblock {\em arXiv preprint arXiv:1810.06546}, 2018.

\bibitem{trinh2021interpretable}
Loc Trinh, Michael Tsang, Sirisha Rambhatla, and Yan Liu.
\newblock Interpretable and trustworthy deepfake detection via dynamic
  prototypes.
\newblock In {\em Proceedings of the IEEE/CVF winter conference on applications
  of computer vision}, pages 1973--1983, 2021.

\bibitem{uchiyama2022visually}
Tomoki Uchiyama, Naoya Sogi, Koichiro Niinuma, and Kazuhiro Fukui.
\newblock Visually explaining 3d-cnn predictions for video classification with
  an adaptive occlusion sensitivity analysis.
\newblock {\em arXiv preprint arXiv:2207.12859}, 2022.

\bibitem{ungar2007hyperbolic}
Abraham~A Ungar.
\newblock The hyperbolic square and mobius transformations.
\newblock {\em Banach Journal of Mathematical Analysis}, 2007.

\bibitem{wah2011caltech}
Catherine Wah, Steve Branson, Peter Welinder, Pietro Perona, and Serge
  Belongie.
\newblock The caltech-ucsd birds-200-2011 dataset.
\newblock 2011.

\bibitem{wang2022hint}
Andong Wang, Wei-Ning Lee, and Xiaojuan Qi.
\newblock Hint: Hierarchical neuron concept explainer.
\newblock In {\em Proceedings of the IEEE/CVF Conference on Computer Vision and
  Pattern Recognition}, pages 10254--10264, 2022.

\bibitem{wang2021interpretable}
Jiaqi Wang, Huafeng Liu, Xinyue Wang, and Liping Jing.
\newblock Interpretable image recognition by constructing transparent embedding
  space.
\newblock In {\em Proceedings of the IEEE/CVF International Conference on
  Computer Vision}, pages 895--904, 2021.

\bibitem{wang2016temporal}
Limin Wang, Yuanjun Xiong, Zhe Wang, Yu Qiao, Dahua Lin, Xiaoou Tang, and
  Luc~Van Gool.
\newblock Temporal segment networks: Towards good practices for deep action
  recognition.
\newblock In {\em European conference on computer vision}, pages 20--36.
  Springer, 2016.

\bibitem{warrington1975selective}
Elizabeth~K Warrington.
\newblock The selective impairment of semantic memory.
\newblock {\em The Quarterly journal of experimental psychology},
  27(4):635--657, 1975.

\bibitem{yuksekgonul2022post}
Mert Yuksekgonul, Maggie Wang, and James Zou.
\newblock Post-hoc concept bottleneck models.
\newblock {\em arXiv preprint arXiv:2205.15480}, 2022.

\bibitem{zhou2016learning}
Bolei Zhou, Aditya Khosla, Agata Lapedriza, Aude Oliva, and Antonio Torralba.
\newblock Learning deep features for discriminative localization.
\newblock In {\em Proceedings of the IEEE conference on computer vision and
  pattern recognition}, pages 2921--2929, 2016.

\bibitem{zhu2020hypertext}
Yudong Zhu, Di Zhou, Jinghui Xiao, Xin Jiang, Xiao Chen, and Qun Liu.
\newblock Hypertext: Endowing fasttext with hyperbolic geometry.
\newblock {\em arXiv preprint arXiv:2010.16143}, 2020.

\end{thebibliography}
}
\clearpage

\section{Supplementary Materials}
\subsection{Hierarchical Action Embeddings} 
\label{sec:hier_emb}
Incorporating the prior hierarchical knowledge about actions into the network requires that we represent them as embeddings. 
In this section we detail how to learn those action embeddings in hyperbolic space, in the next section, we explain how to learn hierarchical prototypes that are optimized by aligning them to the action embeddings in hyperbolic space. 
Given the set of action classes $\mathcal{A}=\{1,2,...,|\mathcal{A}|\}$, in hierarchical action recognition we also consider their ancestor classes $\mathcal{H}=\{|\mathcal{A}|+1,|\mathcal{A}|+2,...,|\mathcal{A}|+|\mathcal{H}|\}$, which allows us to construct a hierarchical tree with three levels, i.e., grandparent, parent, and child (see Figure 2 right in the paper). This process of embedding the hierarchies is performed once, offline, per dataset. However, this process can easily be repeated for alternative hierarchies.

\myparagraph{Learning Action Embeddings.}
We map the action hierarchy $\mathcal{A}\cup \mathcal{H}$ into the shared hyperbolic space $\DD^n$ to obtain hierarchical action embeddings, which are used as action class templates in the next section. Let $\mathcal{P}=\{(u,v)|u=\phi(v)\}$ be the positive pair of $v$ and its parent $\phi(v)$ and $\mathcal{N}=\{(u',v')|u'\neq \phi(v')\}$ be the negative pairs. The discriminative loss akin to \cite{long2020searching}:
\begin{equation}
\label{eq3.1.1}
\CL(\CP, \CN, \mathbf{\Phi}) = \CL_{H}(\CP, \CN) + \lambda \cdot \CL_S(\mathbf{\Phi}),
\end{equation}
where $\mathbf{\Phi}$ stands for class templates matrix and its $c-th$ column $\mathbf{\Phi}_c$ is the template vector of class $c$ in $\DD^n$. For the loss function, $\mathcal{L}_{H}$ encourages the preservation of parent-child relations and $\mathcal{L}_{S}$ enforces separation among different sub-hierarchies. The first part  $\CL_{H}$ is akin to \cite{Nickel:2017wz}, where the wrongly positioned child-parent pairs will be penalized:
\begin{equation}
\mathcal{L}_{H}(\CP, \CN) = \ \sum_{(\vu, \vv) \in \CP} \log \bigg(\frac{e^{-d(\vu, \vv)}}{\sum_{(\vu,\vv^{\prime}) \in \CN} e^{-d\left(\vu, \vv^{\prime}\right)}} \bigg),
\end{equation}
where $-d(\vu, \vv)$ is the hyperbolic distance between two action embeddings $\vu$ and $\vv$, which can be written in short-hand notation:
\begin{equation}
\label{eq3.1.2}
d(\vv, \vu) :=2 \operatorname{arctanh}\left(\left\|-\vv \oplus \vu \right\|\right),
\end{equation}
where $\oplus$ indicates the Möbius addition~\cite{ungar2007hyperbolic} in $1-$curved hyperbolic space $\DD^n$.

In the second part, we encourage the separation among sibling relationships, where we update $\vPhi$ with separation loss:
\begin{equation}
\label{eq:3.1.4}
\mathcal{L}_{S}(\vPhi) = -\sum_{i \in |\CA| }||\tilde{\vPhi}^T_i\tilde{\vPhi}_i||_F + \gamma ||(\hat{\vPhi_i} \hat{\vPhi_i}^{T}-\mathbf{I})||_F,
\vspace{-2mm}
\end{equation}
where $\hat{\vPhi}$ consists of the non-sibling vectors with respect to action class $i$ while $\tilde\vPhi$ consists of $i$'s sibling vectors. 

After learning with the above objectives, we obtain $\vPhi$, a matrix of action template vectors including both actions $\CA$ and ancestor (parent and grandparent) actions $\CH$.
\subsection{Hierarchies for UCF-101 and ActivityNet}
\begin{figure*}
    \centering
    \includegraphics[width=\linewidth]{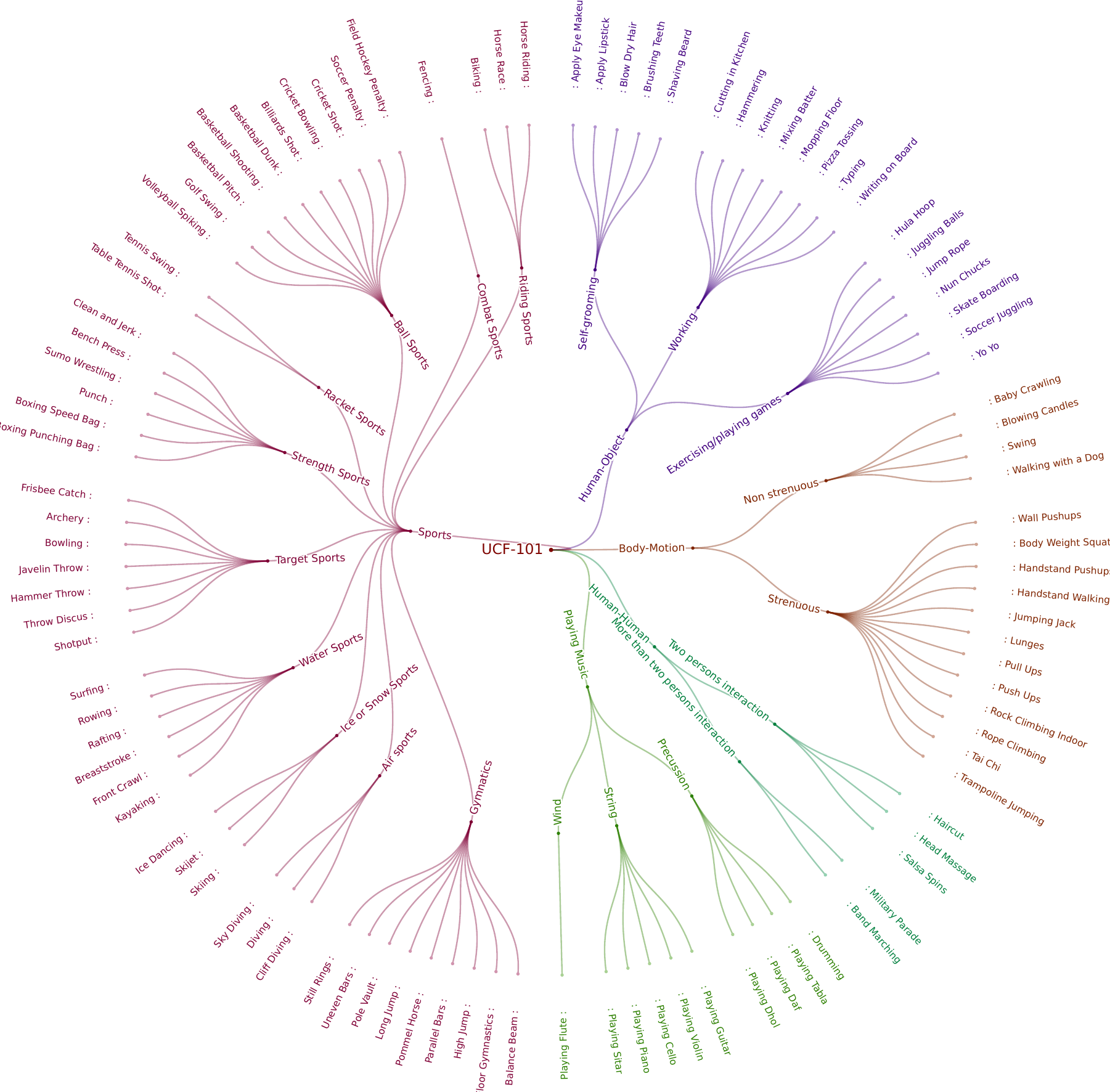}
    \caption{\textbf{Hierarchy for UCF-101.} Schematic representation of the hierarchy defined for UCF-101 dataset. The three levels of hierarchy are grand parent (\textit{playing music}), parent (\textit{wind, string, percussion}) and children (\textit{playing flute, playing guitar, drumming} etc) classes.}
    \label{fig:hierarchy-1}
\end{figure*}

\begin{figure*}
    \centering
    \includegraphics[width=\linewidth]{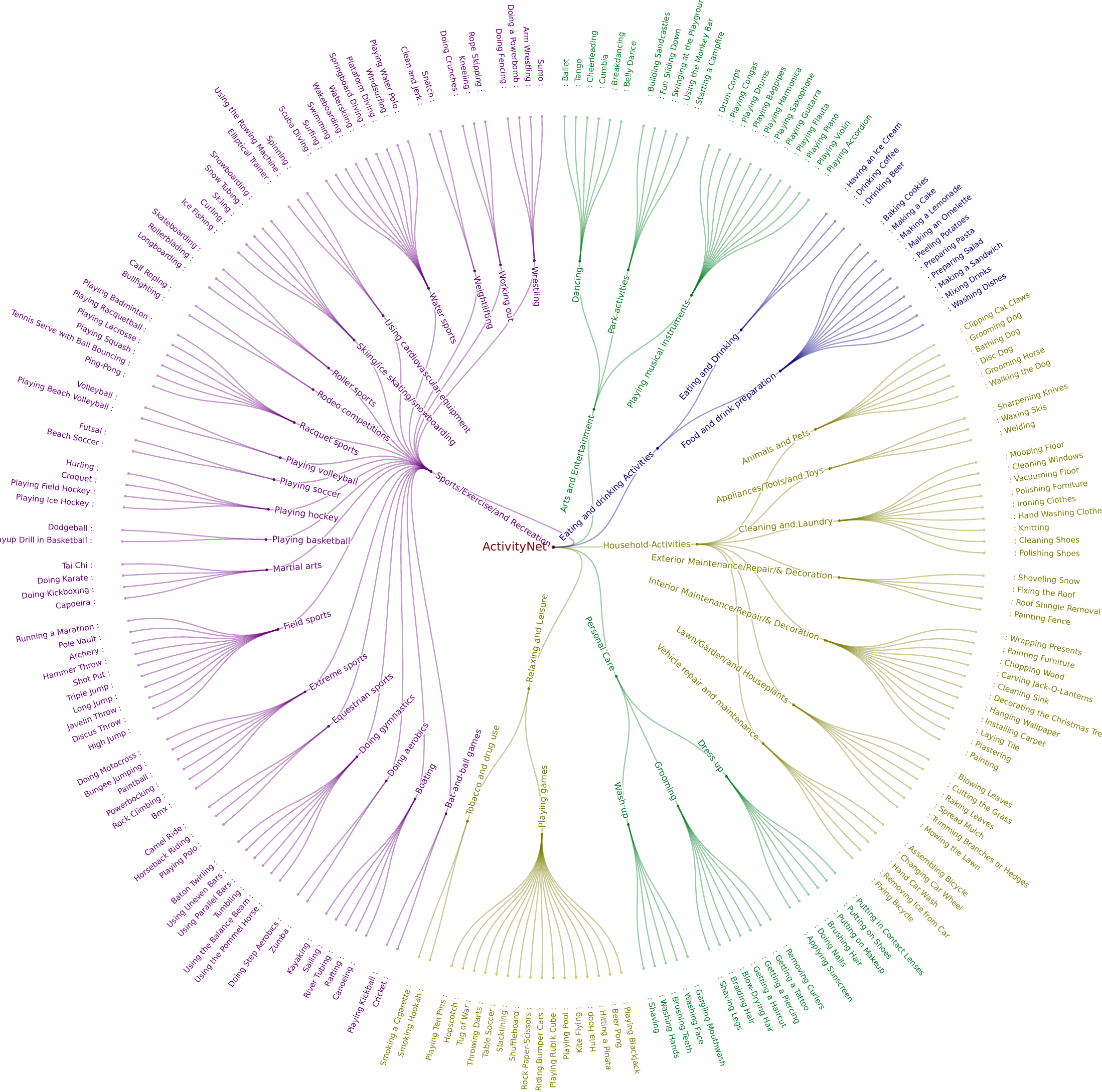}
    \caption{\textbf{Hierarchy for ActivityNet.} Schematic representation of the hierarchy defined for Activity dataset. The three levels of hierarchy are grand parent (\textit{personal care}), parent (\textit{dress up, grooming, wash up}) and children (\textit{putting on shoes, getting a haircut, shaving} etc) classes. }
    \label{fig:hierarchy-2}
\end{figure*}

Figure \ref{fig:hierarchy-1} and \ref{fig:hierarchy-2} show the hierarchies for UCF-101 \cite{soomro2012ucf101} and ActivityNet \cite{caba2015activitynet} respectively. We define three levels of hierarchy with the number of classes at level one, two, and three being 5, 20, and 101 respectively for UCF-101. For example, one of the grand parent class is \textit{playing music}, parents are \textit{wind, string, percussion} and children are \textit{playing flute, playing guitar, drumming} and more. The classes at the third level (i.e., child level) of the hierarchy are the 101 original classes of the dataset. 

ActivityNet contains 200, 38, and 6 classes in level one, two, and three respectively (see Figure \ref{fig:hierarchy-2}). For instance, one of the grand parent is \textit{personal care} and the parents are \textit{dress up, grooming, wash up} and children are \textit{putting on shoes, getting a haircut, shaving} etc. 
\subsection{Visual Explanations}
\begin{figure*}
    \centering
    \includegraphics[width=\linewidth]{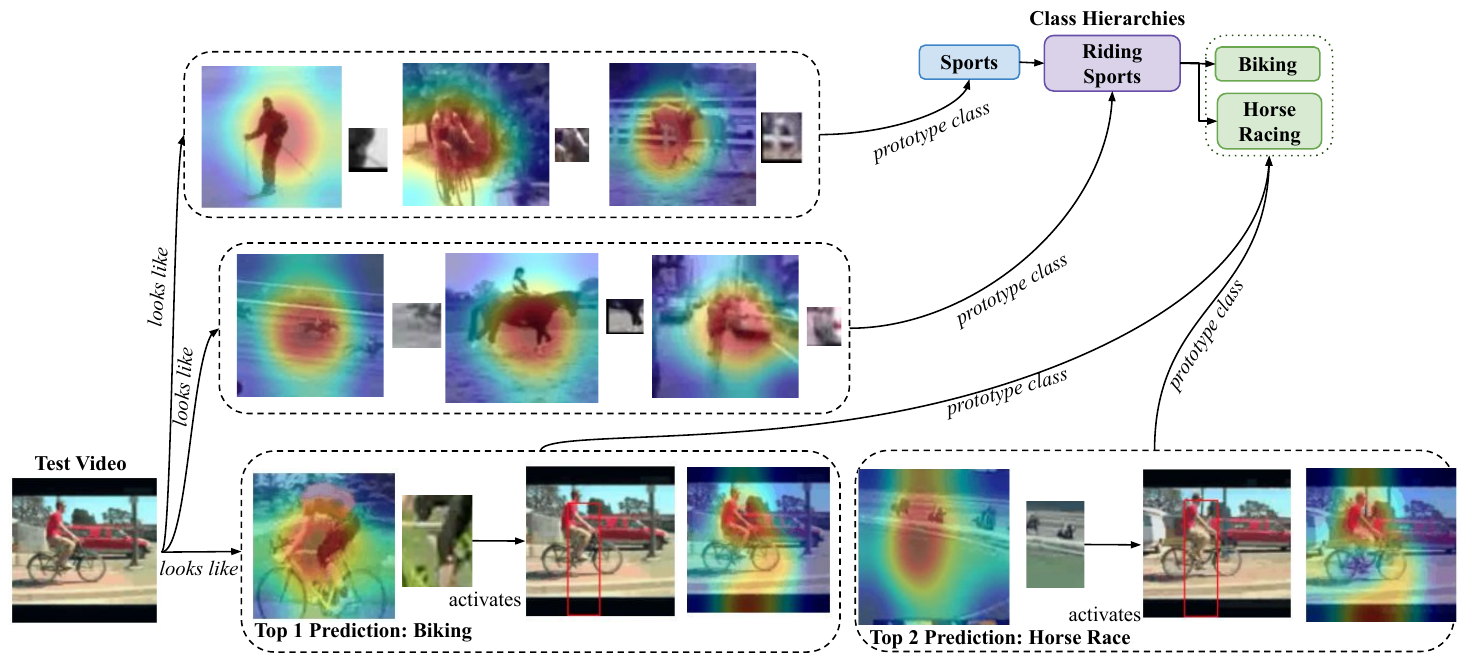}
    \caption{\textbf{Hierarchical Explanations.} This example shows the prototypes from grandparent \textit{sports} class, parent \textit{riding sports} class and ground truth \textit{biking} class, we also observe that the top 2 prediction for the model is its sibling \textit{horse race} class. This conforms that our model is learning hierarchical relations between classes.}
    \label{fig:supp-2}
\end{figure*}

\begin{figure*}
    \centering
    \includegraphics[width=\linewidth]{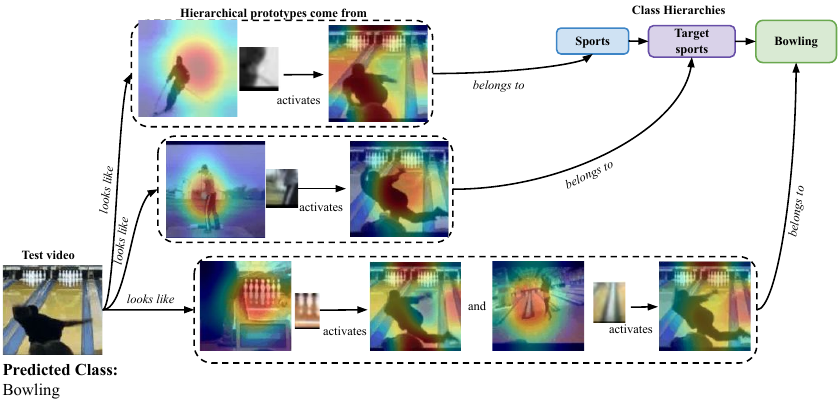}
    \caption{\textbf{Hierarchical Explanations.} Schematic representation of the hierarchical prototype-based reasoning process of our proposed Hierarchical Prototype Explainer.}
    \label{fig:supp-1}
\end{figure*}

\begin{figure*}
    \centering
    \includegraphics[width=\linewidth]{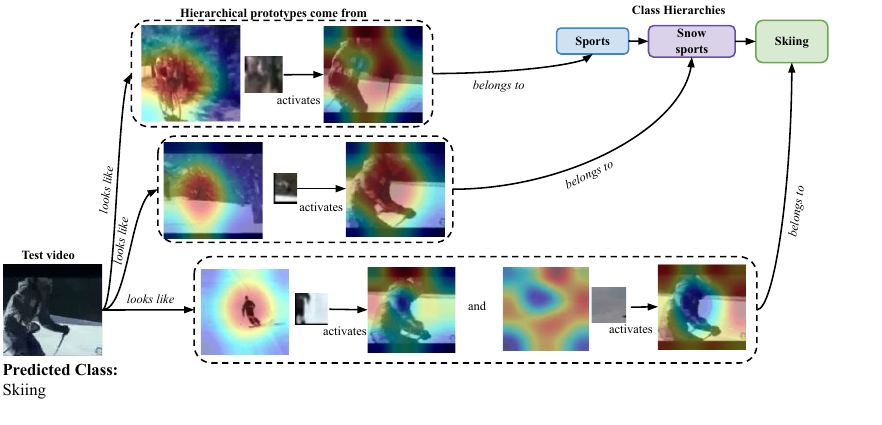}
    \caption{\textbf{Hierarchical Explanations.} Schematic representation of the hierarchical prototype-based reasoning process of our proposed Hierarchical Prototype Explainer.}
    \label{fig:supp-4}
\end{figure*}

\myparagraph{Hierarchical Explanations.}
Figure \ref{fig:supp-2}, \ref{fig:supp-1} and \ref{fig:supp-4} show the quantitative results for our multi-level explanations.

\begin{figure*}
    \centering
    \includegraphics[width=\linewidth]{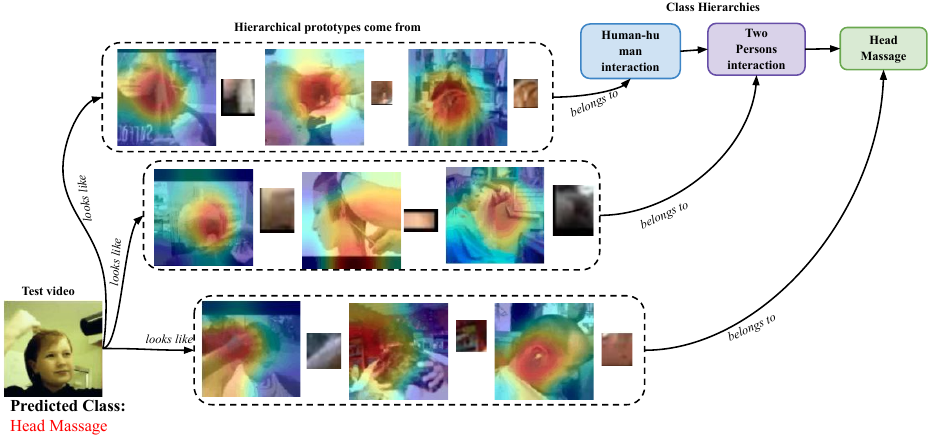}
    \caption{\textbf{Effectiveness in case of failure.} Our multi-level explanations provide useful information even in the case of misclassification through the prototypes learned for parent and grandparent classes.  }
    \label{fig:supp-3}
\end{figure*}

\begin{figure*}
    \centering
    \includegraphics[width=\linewidth]{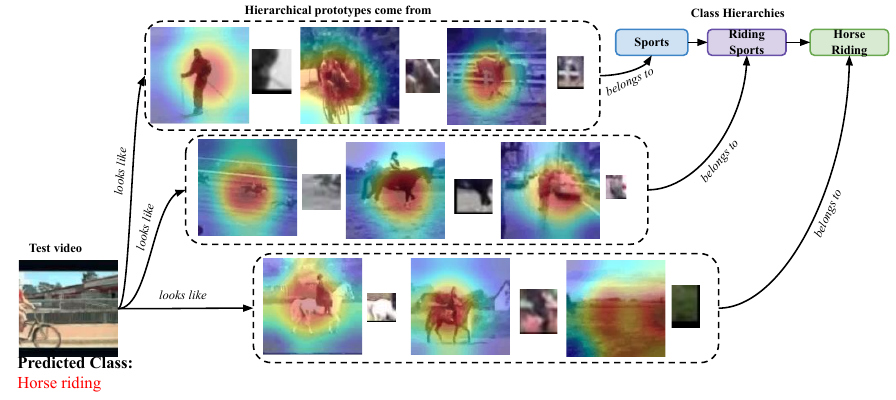}
    \caption{\textbf{Effectiveness in case of failure.} Our multi-level explanations provide useful information even in the case of misclassification through the prototypes learned for parent and grandparent classes. }
    \label{fig:supp-5}
\end{figure*}

\myparagraph{Effectiveness in Case of Failure.}
 Figure \ref{fig:supp-3} and \ref{fig:supp-5} show the effectiveness of our method in case of failure.


\end{document}